# A global optimization SAR image segmentation model can be easily transformed to a general ROF denoising model

Guangming liu
Yantai Science and Technology Association,email:1733823365@qq.com

**Abstract**—In this paper, we propose a novel locally statistical active contour model (LACM) based on Aubert-Aujol (AA) denoising model and variational level set method, which can be used for SAR images segmentation with intensity inhomogeneity. Then we transform the proposed model into a global optimization model by using convex relaxation technique. Firstly, we apply the Split Bregman technique to transform the global optimization model into two alternating optimization processes of Shrink operator and Laplace operator, which is called SB_LACM model. Moreover, we propose two fast models to solve the global optimization model , which are more efficient than the SB_LACM model. The first model is: we add the proximal function to transform the global optimization model to a general ROF model[29], which can be solved by a fast denoising algorithm proposed by R.-Q.Jia, and H.Zhao; [29] was submitted on 29-Aug-2013, and our early edition was ever submitted to TGRS on 12-Jun-2012, Venkatakrishnan et al. [30] proposed their "**PnP algorithm**" on 29-May-2013, so Venkatakrishnan and we proposed the "**PnP algorithm**" almost simultaneously. Thus we obtain a fast segmentation algorithm with global optimization solver that does not involve partial differential equations or difference equation, and only need simple difference computation. The second model is: we use a different splitting approach than one model to transform the global optimization model into a differentiable term and a general ROF model term, which can be solved by the same technique as the first model.

Experiments using some challenging synthetic images and Envisat SAR images demonstrate the superiority of our proposed models with respect to the state-of-the-art models.

## Ⅰ.INTRODUCTION

Segmentation is a crucial step synthetic aperture radar (SAR) images' automatic interpretation. Due to the presence of speckle be modeled as strong multiplicative noise, segmentation of SAR images is generally acknowledged as a difficult problem. To cope with the influence of speckle noise on image segmentation, a large number of methods have been proposed. Recently, level set methods (LSMs) have been extensively applied to SAR images segmentation[1]-[5]. We know that level set methods can be divided into two major categories: region-based models [1-9] and edge-based models [10-13]. In order to make use of edge information and region characteristic, region-based model is often integrated with edge-based model to form the energy functional of image segmentation models. There are some advantages of level set methods over classical image segmentation methods, such as edge detection, thresholding, and region grow: level set methods can give sub-pixel accuracy of object boundaries, and can be easily formulated under a energy functional minimization, and can provide smooth and closed contour as a result of segmentation and so forth. However, a common difficulty with above variational image segmentation models [1-13] is that the energy functionals to be minimized are all not convex. Thus we may obtain a local minimizer rather than a unique global minimizer from these non-convex energy functionals. This is a more serious problem because local minima of image segmentation often has completely wrong levels of detail and scale. In order to resolve the problems associated with non-convex models, Chan-Esedoglu-Nikolova [14] proposed a convex relaxation approach for image segmentation models. Bresson et al. [15] use a splitting/regularization approach to minimize the segmentation energy functional, However, this scheme lows down as the accuracy increases. In addition, the globally convex segmentation models [14-15] all contain a total variation (TV) term that is not differentiable in zero, making them difficult to compute. It is well known that the split Bregman method is a technique for fast minimization of L1 regularized energy functional. Goldstein-Osher (GO) [16-17] proposed a more efficient way to compute the TV term by applying the split Bregman method.

On the one hand, we know SAR images are affected by speckle, a multiplicative gamma noise that gives the images a grainy appearance and makes the interpretation of SAR images a more difficult task to achieve. However, Aubert and Aujol (AA) [18] proposed a non-convex energy functional to remove multiplicative gamma noise based on the maximum a posteriori (MAP) regularization approach. We propose a novel local active contour (LACM) model for SAR images segmentation with intensity inhomogeneity based on AA model in this paper. By convex relaxation [14] method and split Bregman technique [17], we can transform the proposed LACM model into a globally convex segmentation model, which only involves two alternating optimization processes of shrink operator and Laplace operator.

On the other hand, it is well known that the ROF denoising model is strictly convex, which always admits a unique solution. By adding the proximal term, we can transform the proposed LACM model into the ROF denoising model. Hence, a variety of numerical schemes have been already proposed in literature in order to solve this problem. Recently, Jia-Zhao[19-20] give a algorithm for the solution of ROF denoising model [21] based on anisotropic TV term, which do not involve partial differential equations, hence is much faster than GO algorithm. Then we further propose a fixed point algorithm for LACW model.

This paper is organized as follows. Section Ⅱ describes the proposed local active contour (LACM) model. Section Ⅲ describes the proposed two fast globally convex segmentation algorithms, and Section Ⅳ describes the experimental results. Section Ⅴ concludes this paper.

## Ⅱ. A NOVEL LOCALLY STATISTICAL ACTIVE CONTOUR MODEL BASED ON AA MODEL AND VARIATIONAL LEVEL SET METHOD

We denote by $R^2$ the usual 2-dimensional Euclidean space $H$. We use $\langle .,. \rangle$ and $\| \|$, respectively, to denote the inner product and the corresponding L2 norm of an Euclidean space $H$ while $\| \|_1$ is used to denote the L1 norm. We denote by $\nabla_x^T$ ($\nabla_y^T$) the conjugate of the gradient operator $\nabla_x$ ($\nabla_y$).

**A. AA model**

Given a observed intensity image $f:\Omega \to R(f>0)$, where $\Omega$ is a bounded open subset of $R^2$, speckle is well modeled as a multiplicative random noise $n$, which is independent of the true image $u$, i.e. $f = u \cdot n$. We know that fully developed multiplicative speckle noise is Gamma distributed with mean value $\mu_n = 1$ and variance $\sigma_n{}^2 = 1/L$, where $L$ is the equivalent number of independent looks of the image.

Aubert and Aujol (AA) [18] proposed a non-convex energy functional to remove multiplicative gamma noise based on the maximum a posteriori (MAP) regularization approach. The energy functional of AA model can be formulated as:

$$E(u) = \theta \int_\Omega |\nabla u| dx + \mu \int_\Omega (\log u + \frac{f}{u}) dx \qquad (1)$$

The first term of (1) is the TV regularization term, and the second term is the data fitting term, and $\mu$ and $\theta$ are positive constant parameter to balance the first term and second term. The distinctive feature of the TV regularization term and its various variants is that edges of images are preserved in the denoised image.

**B.LACM model**

We adopt the idea of Chan-Vese (CV) model [6] on the two-phase segmentation question, and suppose that the true image $u$ is piecewise constant, i.e. $u = C_1$ for $x \in \Omega_1$, and $u = C_2$ for $x \in \Omega_2$, where $\{\Omega_1, \Omega_2\}$ is a partitioning of the image domain $\Omega$. Equation (1) can be written in the level set formulation by adopting the idea of the CV model as:

$$E(\phi, C_1, C_2) = \theta \int_\Omega \delta(\phi) |\nabla \phi| dx + \mu \sum_{i=1}^{2} \int_\Omega (\log C_i + \frac{f}{C_i}) M_i(\phi) dx \qquad (2)$$

Where $\phi$ is a level set function, $M_1 = H(\phi)$, $M_2 = 1 - H(\phi)$, $H(\phi)$ is the Heaviside function. Then using the level set formulation, the true image $u$ can be expressed:

$$u = C_1 H(\phi) + C_2 (1 - H(\phi)) . \qquad (3)$$

$H(\phi)$ is often approximated by a smooth function $H_\varepsilon(\phi)$ defined by

$$H_\varepsilon(\phi) = \frac{1}{2}[1 + \frac{2}{\pi} \arctan(\frac{\phi}{\varepsilon})] \qquad (4)$$

to automatically detect interior contours and insure the computation of a global minimizer in this paper.

The derivative of $H_\varepsilon(\phi)$ is also approximated by a smooth function

$$\delta_\varepsilon(\phi) = \frac{\partial H_\varepsilon(\phi)}{\partial \phi} = \frac{1}{\pi}(\frac{\varepsilon}{\varepsilon^2 + \phi^2}) \qquad (5)$$

Therefore, the energy functional (1) becomes:

$$E_\varepsilon(\phi, C_1, C_2) = \theta \int_\Omega \delta_\varepsilon(\phi) |\nabla \phi| dx + \mu \sum_{i=1}^{2} \int_\Omega (\log C_i + \frac{f}{C_i}) M_i^\varepsilon(\phi) dx \qquad (6)$$

Where $M_1^\varepsilon = H_\varepsilon(\phi)$, $M_2^\varepsilon = 1 - H_\varepsilon(\phi)$.

We further modify (6) to incorporate information from an edge detector $g$, which can make (6) more likely to favor segmentation along curves where the edge detector function $g$ is minima. On the other hand, in order to penalize the deviation of $\phi$ from a signed distance function during its evolution, we add a level set regularization term [13] to equation (6). Thus we give a hybrid model between region-based method and edge-based method, which can be used to segment image with intensity homogeneity:

$$
\begin{aligned}
& E_{\varepsilon,g}(\phi, C_1, C_2) \\
& = \theta \int_{\Omega} g \delta_{\varepsilon}(\phi) |\nabla \phi| dx + \mu \sum_{i=1}^{2} \int_{\Omega} (\log C_i + \frac{f}{C_i}) M_i^{\varepsilon}(\phi) dx + v \cdot \int_{\Omega} \frac{1}{2} (|\nabla \phi(x)| - 1)^2 dx
\end{aligned} \tag{7}
$$

Where $v$ is a tune parameter ; $g$ is a positive and decreasing edge detector function which is often defined as $g = \frac{1}{1 + \beta |\nabla f_{\sigma} \otimes I|^2}$, and $g$ usually takes smaller values at object boundaries than at other locations; The parameter $\beta$ controls the details of the segmentation, and $\nabla f_{\sigma} \otimes I$ is used to smooth the image to reduce the noise; In order to efficiently smooth multiplicative noise, we use the infinite symmetric exponential filter (ISEF) $f_{\sigma}(x) = \frac{1}{2\sigma} e^{-\frac{|x|}{\sigma}}$ with standard deviation $\sigma = 1.2$ and of size $15 \times 15$ in edge detector $g$, which is optimal in the case of multiplicative noise [22]; And ISEF has better edge localization precision than other edge detectors [23].

According to (7) and region-scalable-fitting (RSF) model [8], we then propose a novel local active contour (LACM) model for SAR images segmentation with intensity inhomogeneity. The LACM model can be written in the level set formulation as:

$$
\begin{aligned}
& E_{\varepsilon,g,K_{\sigma}}^{LACM}(\phi, C_1, C_2) \\
& = \theta \int_{\Omega} g \delta_{\varepsilon}(\phi) |\nabla \phi| dx + \mu \sum_{i=1}^{2} \int_{\Omega} \left( \int_{\Omega} K_{\sigma}(x-y) (\log C_i(x) + \frac{f(y)}{C_i(x)}) M_i^{\varepsilon}(\phi(y)) dy \right) dx \\
& + v \cdot \int_{\Omega} \frac{1}{2} (|\nabla \phi(x)| - 1)^2 dx
\end{aligned} \tag{8}
$$

Where $K_{\sigma}$ is a Gaussian kernel with standard deviation $\sigma$.

For fixed level set function $\phi$, we minimize the function $E^{LACM}_{\varepsilon,g,K_\sigma}(\phi, C_1, C_2)$ with respect to the constant $C_1(x)$ and $C_2(x)$. By calculus of variations, it is also easy to solve them by

$$C_1(x) = \frac{K_\sigma(x) * (H_\varepsilon(\phi(x)) f(x))}{K_\sigma(x) * H_\varepsilon(\phi(x))}, \quad C_2(x) = \frac{K_\sigma(x) * ((1 - H_\varepsilon(\phi(x))) f(x))}{K_\sigma(x) * (1 - H_\varepsilon(\phi(x)))} \tag{9}$$

Minimization of the energy functional $E^{LACM}_{\varepsilon,g,K_\sigma}(\phi, C_1, C_2)$ with respect to $\phi$, can be obtained by solving gradient descent flow equation:

$$\frac{\partial \phi}{\partial t} = \delta_\varepsilon(\phi)(\theta div(g \frac{\nabla \phi}{|\nabla \phi|}) - \mu \cdot \eta) + v \cdot (\nabla^2 \phi - div(\frac{\nabla \phi}{|\nabla \phi|})) \tag{10}$$

Where,

$$\eta = \int_\Omega K_\sigma(x-y)(\log C_1(y) + \frac{f(x)}{C_1(y)}) dy - \int_\Omega K_\sigma(x-y)(\log C_2(y) + \frac{f(x)}{C_2(y)}) dy .$$

## Ⅲ. GLOBALLY CONVEX SEGMENTATION ALGORITHMS BASED ON LACM MODEL

We drop the last term of equation (10), which regularized the level set function to be close to a distance function and set $\theta = 1$. We can get a simplified flow equation based on the globally convex segmentation (GCS) [14][15] and its application in [16-17], which has the coincident stationary solution with (10) as follows:

$$\frac{\partial \phi}{\partial t} = div(g \frac{\nabla \phi}{|\nabla \phi|}) - \mu \cdot \eta \tag{11}$$

The simplied flow represent gradient descent for minimizing the energy functional:

$$E(\phi) = \|\nabla \phi\|_g + \mu < \phi, \eta > \tag{12}$$

Where $\|\nabla \phi\|_g = \int_\Omega g |\nabla \phi| dx$, $< \phi, \eta > = \int_\Omega \phi \cdot \eta dx$.

In order to guarantee the unique global minimizer of energy functional (12), we restrict the solution $\phi$ to lie in a finite interval ($0 \le \phi \le 1$).We suppose the weighted

TV term of the energy functionals (12) is anisotropic. By applying the anisotropic TV term to (12), we can obtain a globally convex segmentation model as follows:

$$E(\phi)=\min_{0\le\phi\le1}\left\|\nabla_x\phi\right\|_g+\left\|\nabla_y\phi\right\|_g+\mu<\phi,\eta> \tag{13}$$

**A. Split Bregman method for LACM (SB_LACM)model**

In the past, solutions of the TV model were based on nonlinear partial differential equations and the resulting algorithms were very complicated. It is well known that the split Bregman method is a technique for fast minimization of L1 regularized energy functional. A break through was made by Goldstein-Osher (GO) [16-17].They proposed a more efficient way to compute the TV term based on the split Bregman method. To apply the split Bregman method to (15), we introduce auxiliary variables $d_x\leftarrow\nabla_x\phi$, $d_y\leftarrow\nabla_y\phi$, and add a quadratic penalty function to weakly enforce the resulting equality constraint which results in the following unconstrained problem:

$$\begin{aligned}&(\phi^{k+1},d_x^{k+1},d_y^{k+1})=\arg\min_{0\le\phi\le1}\left\|d_x\right\|_g+\left\|d_y\right\|_g+\mu<\phi,\eta>+\\&\frac{\lambda}{2}\left\|d_x-\nabla_x\phi\right\|^2+\frac{\lambda}{2}\left\|d_y-\nabla_y\phi\right\|^2\end{aligned} \tag{14}$$

We then apply split Bregman method to strictly enforce the constraints $d_x=\nabla_x\phi$, $d_y=\nabla_y\phi$. The resulting optimization problem becomes:

$$\begin{aligned}&(\phi^{k+1},d_x^{k+1},d_y^{k+1})=\arg\min_{0\le\phi\le1}\left\|d_x\right\|_g+\left\|d_y\right\|_g+\mu<\phi,\eta>+\\&\frac{\lambda}{2}\left\|d_x-\nabla_x\phi-b_x^k\right\|_2^2+\frac{\lambda}{2}\left\|d_y-\nabla_y\phi-b_y^k\right\|_2^2\end{aligned} \tag{15}$$

$$b_x^{k+1}=b_x^k+\nabla_x\phi^{k+1}-d_x^{k+1},b_y^{k+1}=b_x^k+\nabla_y\phi^{k+1}-d_y^{k+1} \tag{16}$$

For fixed $\vec{d}$, the Euler-Lagrange equation of optimization problem (15) with respect to $\phi$ is:

$$\Delta\phi^{k+1}=\frac{\mu\eta}{\lambda}-\nabla_x^T(d_x^k-b_x^k)-\nabla_y^T(d_y^k-b_y^k) \tag{17}$$

For fixed $\phi$, minimization of (15) with respect to $\vec{d}$ gives:

$$d_x^{k+1} = shrink_{g/\lambda}(\nabla_x \phi^{k+1} + b_x^k), d_y^{k+1} = shrink_{g/\lambda}(\nabla_x \phi^{k+1} + b_y^k) \quad (18)$$

Where $shink_{g/\lambda}(x) = \text{sgn}(x)\max\{|x| - g/\lambda, 0\}$.

By using central discretization for Laplace operator and backward difference for divergence operator, the numerical scheme for (17) becomes:

$$\alpha_{i,j} = d_{i-1,j}^x - d_{i,j}^x - b_{i-1,j}^x + b_{i,j}^x + d_{i,j-1}^y - d_{i,j}^y - b_{i,j-1}^y + b_{i,j}^y \quad (19)$$

$$\beta_{i,j} = \frac{1}{4}(\phi_{i-1,j} + \phi_{i+1,j} + \phi_{i,j-1} + \phi_{i,j+1} - \frac{\mu \cdot \eta^k}{\lambda} + \alpha_{i,j}) \quad (20)$$

$$\phi_{i,j} = \max\{\min\{\beta_{i,j}, 1\}, 0\} \quad (21)$$

As the optimal $\phi$ is found, the segmented region can be found by thresholding the level set function $\phi(x)$ for some $\gamma \in (0,1)$: $\Omega_1 = \{x : \phi(x) > \gamma\}$.

The split Bregman method for the minimization problem (15) can be summarized as follows: **SB_LACM model:**

Given: noisy image $f$; $\lambda > 0$, $\mu > 0$

Initialization: $b^0 = 0, d^0 = 0$, $\phi^0 = f / \max(f)$

1.**while** $\|\phi^{k+1} - \phi^k\| \geq vol$ **do**

2.Define

$$\eta = \int_\Omega K_\sigma(x-y)(\log C_1(y) + \frac{f(x)}{C_1(y)})dy - \int_\Omega K_\sigma(x-y)(\log C_2(y) + \frac{f(x)}{C_1(y)})dy$$

3. Compute (19)-(21).

4. Compute (18).

5.Compute (16)

6.Find $\Omega_1 = \{x : \phi(x) > \gamma\}$

7.Update $C_1^{k+1}, C_2^{k+1}$

8.**end while**

**END**

**B. Fixed point algorithm 1 for LACM (FP1_LACM)model**

We know the energy functional of image segmentation (13) does not have a unique global minimizer because it is homogeneous degree one. However, the ROF denoising model always admits a unique solution because the energy functional is strictly convex. In order to utilizing the convexity of ROF denoising model and favor segmentation along curves where the edge detector function is minima, we give a discrete version weighted ROF (WROF) denoising model based on anisotropic TV term as follows:

$$E_{WROF}(\phi)=\|\nabla_x\phi\|_g+\|\nabla_y\phi\|_g+\frac{\alpha}{2}\|\phi-f\|^2 \tag{22}$$

Where $\alpha>0$ is an appropriately chosen positive parameter.

Note that GO algorithm [16-17] still requires solving a partial difference equation in each iteration step. Recently, Jia-Zhao (JZ)[19-20] proposed a fast algorithm for image denoising based on TV term, which is very simple and does not involve partial differential equations or difference equation. We further transform the proposed model into a generalized ROF model by adding a proximity term ,and it can be solved by a fast denoising algorithm proposed by Jia-Zhao or soved by BM3D and NLM denoising algorithm, which also provide a unified solution framework for formally generalized-ROF-like subproblems arising in multivariate splitting algorithms.

In order to apply the JZ algorithm to image segmentation question (13), we first suppose that $\phi^k$ and $\eta^k$ are known and reformulate (13) by adding a proximity term $\frac{\alpha}{2}\|\phi-\phi^k\|^2$ as:

$$\begin{aligned}\phi^{k+1}&=\arg\min_{0\le\phi\le 1}\|\nabla_x\phi\|_g+\|\nabla_y\phi\|_g+\mu<\phi,\eta>+\frac{\alpha}{2}\|\phi-\phi^k\|^2\\&=\arg\min_{0\le\phi\le 1}\|\nabla_x\phi\|_g+\|\nabla_y\phi\|_g+\mu<\phi-\phi^k,\eta>+\frac{\alpha}{2}\|\phi-\phi^k\|^2\\&=\arg\min_{0\le\phi\le 1}\|\nabla_x\phi\|_g+\|\nabla_y\phi\|_g+\frac{\alpha}{2}\left\|\phi-(\phi^k-\frac{\mu\eta}{\alpha})\right\|^2\end{aligned} \tag{23}$$

Treating $\phi^k-\dfrac{\mu\eta}{\alpha}$ as the **“noisy image”**, (23) minimizes the residue between the **“clean image”** $\phi$ and $\phi^k-\dfrac{\mu\eta}{\alpha}$ using the total variation norm, then (23) is the total variation denoising problem. So we propose **FP1_LACM model** based on JZ algorithm to solve (23) as follows:

$$b_x^k = (I - shrink_{1/\lambda})(\nabla_x \phi^k + b_x^{k-1}), b_y^k = (I - shrink_{1/\lambda})(\nabla_y \phi^k + b_y^{k-1}) \quad (24)$$

$$\phi^{k+1} = \phi^k - \frac{\mu\eta}{\alpha} - \frac{\lambda}{\alpha}(\nabla_x^T b_x^k + \nabla_y^T b_y^k) \quad (25)$$

$$\phi^{k+1} = \max\{\min\{\phi^{k+1}, 1\}, 0\} \quad (26)$$

According to [24-25], we infer from (24) that operator $(I - shrink_{1/\lambda})(I - \frac{\lambda}{\alpha}\nabla\nabla^T)$ is nonexpansive when $\frac{\lambda}{\alpha}$ is less than $\frac{1}{4}\sin^{-2}\frac{(N-1)\pi}{2N}$ , which is slightly bigger than 1/4 . In order to accelerate the convergence of (24) , we adopt the following iteration scheme by utilizing *k*-averaged operator theory ( see more details in [24]):

$$\begin{aligned} b_x^k &= t \cdot b_x^{k-1} + (1-t)\cdot(I - shrink_{1/\lambda})(\nabla_x \phi^k + b_x^{k-1}) \\ b_y^k &= t \cdot b_y^{k-1} + (1-t)\cdot(I - shrink_{1/\lambda})(\nabla_y \phi^k + b_y^{k-1}) \end{aligned} \quad (27)$$

Where the weight factor $t \in (0,1)$ is called the relaxation parameter. As the optimal $\phi(x)$ is found, the segmented region can be found by thresholding the function $\phi(x)$ for some $0 < \gamma < 1$: $\Omega_1 = \{x : \phi(x) > \gamma\}$ . The FP1_LACM model for the minimization problem (23) can be summarized as follows: **FP1_LACM model:**

Given: noisy image $f$ ; $\lambda > 0$ , $\mu > 0$ , $\alpha > 0$ , $t \in (0,1)$ , $\gamma \in (0,1)$

Initialization: $b_x^0 = 0$ , $b_y^0 = 0$ , $\phi^0 = f / \max(f)$ , $\Omega_0 = \{x : \phi^0 > \gamma\}$ ,

1.**while** $\left\|\phi^{k+1} - \phi^k\right\| \geq vol$ **do**

2.Define

$$\eta = \int_\Omega K_\sigma(x-y)(\log C_1(y) + \frac{f(x)}{C_1(y)})dy - \int_\Omega K_\sigma(x-y)(\log C_2(y) + \frac{f(x)}{C_1(y)})dy$$

3. Compute (27).

4. Compute (25)-(26).

5.Find $\Omega_1 = \{x : \phi(x) > \gamma\}$

6.Update $C_1^{k+1}, C_2^{k+1}$

7.**end while**

END

### C. Fixed point algorithm 2 for LACM(FP2_LACM) model

We can also reformulate (13) by introducing a term $\frac{\alpha}{2}\|\phi-\varphi\|^2$ as:

$$(\varphi^{k+1},\phi^{k+1}) = \arg\min_{\varphi,0\le\phi\le 1} \|\nabla_x\phi\|_g + \|\nabla_y\phi\|_g + \mu <\varphi,\eta> + \frac{\alpha}{2}\|\phi-\varphi\|^2 \qquad (28)$$

We then apply split Bregman method to (28) and propose FP2_LACM model as follows:

$$\varphi^k = \arg\min_{0\le\varphi\le 1} \frac{\alpha}{2}\left\|\phi^k-\varphi-c^{k-1}\right\|^2 + \mu <\varphi,\eta> \qquad (29)$$

$$c^k = c^{k-1} + \varphi^k - \phi^k \qquad (30)$$

$$\phi^{k+1} = \arg\min_{0\le\phi\le 1} \|\nabla_x\phi\|_g + \|\nabla_y\phi\|_g + \frac{\alpha}{2}\left\|\phi-\varphi^k-c^k\right\|^2 \qquad (31)$$

We can get a solution of (29)-(31) as follows:

$$\varphi^k = \phi^k - c^{k-1} - \frac{\mu}{\alpha}\eta \qquad (32)$$

$$\varphi^k = \max\{\min\{\varphi^k,1\},0\} \qquad (33)$$

$$c^k = c^{k-1} + \varphi^k - \phi^k \qquad (34)$$

$$\phi^{k+1} = \varphi^k + c^k - \frac{\lambda}{\alpha}(\nabla_x^T b_x^k + \nabla_y^T b_y^k) \qquad (35)$$

Where $b_x^k$ and $b_y^k$ is also defined as (27). As the optimal $\phi$ is found, the segmented region can be found by thresholding the function $\phi(x)$ for some $\gamma\in(0,1)$ : $\Omega_1=\{x:\phi(x)>\gamma\}$ . The FP2_LACM model for the minimization problem (31) can be summarized as follows: **FP2_LACM model:**

Given: noisy image $f$ ; $\lambda>0$ , $\mu>0$ , $\alpha>0$ , $t\in(0,1)$ , $\gamma\in(0,1)$

Initialization: $b_x^0=0$ , $b_y^0=0$ , $c^0=0$ , $\phi^0=f/\max(f)$

1.**while** $\left\|\phi^{k+1}-\phi^k\right\|\ge vol$ **do**

2.Define

$$\eta=\int_\Omega K_\sigma(x-y)(\log C_1(y)+\frac{f(x)}{C_1(y)})dy-\int_\Omega K_\sigma(x-y)(\log C_2(y)+\frac{f(x)}{C_1(y)})dy$$

3.

Compute (32)-(34).
4. Compute (27).
5. Compute (35).
6.Find $\Omega_1 = \{x : \phi(x) > \gamma\}$
7.Update $C_1^{k+1}, C_2^{k+1}$
8.**end while**
END

## Ⅳ. EXPERIMENTAL RESULTS

We test our model and Algorithms with two synthetic images and two Envisat SAR images in this section. Two synthetic images have the size $125\times125$, and two Envisat SAR images have the size $179\times205$ and $151\times151$, respectively, and the gray-scale in the range between 0 and 255. Two synthetic images were firstly corrupted by shading artifacts, and then were further corrupted Gamma noise with mean value $\mu_n = 1$ and variance ${\sigma_n}^2 = 1/L$ (by setting $L$=1 and $L$=8).

In the following, we first compare our proposed LACM model with the CV model [6], the model [4], the RSF model [8], then compare the SB_LACM model, the FP1_LACM model, and the FP2_LACM model with globally convex segmentation model [5] based on the model [4], and globally convex segmentation model [26] based on the RSF model.

The level set function can be simply initialized as a binary step function which takes a constant value 1 inside a region and another constant value -1 outside.

For synthetic image1 / image 2, we adopt the following parameters.

The parameters of the CV model are chosen as: $\theta = 0.01\times255^2$, $\mu = 1$, $\Delta t = 1$, $\beta = 20$, $\varepsilon = 1$, $v = 0.2$ .The parameters of the model [4] are chosen as: $\theta = 2$, $\mu = 255$, $\Delta t = 1$, $\beta = 20$, $\varepsilon = 1$, $v = 0.2$. The parameters of the RSF model are chosen as $\theta = 200$, $\mu = 0.5, 0.1$, $\Delta t = 1$, $\beta = 20$, $\varepsilon = 1$, $v = 0.2$, $\sigma = 15$ . The parameters of the LACM model are chosen as $\theta = 200, 10$, $\mu = 100$, $\Delta t = 1$,

$\beta = 20$, $\varepsilon = 1$, $v = 0.2$, $\sigma = 15$.

The parameters of the model [5] are chosen as $\lambda = 1000$, $\mu = 0.04\lambda / 0.12\lambda$, $\beta = 20$, $\varepsilon = 1$, $vol = 1$. The parameters of the model [26] are chosen as $\lambda = 1000$, $\mu = 0.0002\lambda / 0.0001\lambda$, $\beta = 20$, $\varepsilon = 1$, $\sigma = 15$, $vol = 1$. The parameters of the SB_LACM model are chosen as $\lambda = 1000$, $\mu = 0.06\lambda / 0.2\lambda$, $\beta = 20$, $\varepsilon = 1$, $\sigma = 15$, $vol = 1$. The parameters of the FP1_LACM model are chosen as $\mu = 2/8$, $\lambda = 1$, $\alpha = 12$, $\beta = 20$, $\varepsilon = 1$, $\sigma = 15$, $t = 1e-5$, $vol = 1$. The parameters of the FP2_LACM model are chosen as $\mu = 1/4$, $\lambda = 1$, $\alpha = 12$, $\beta = 20$, $\varepsilon = 1$ $\sigma = 15$, $t = 1e-5$, $vol = 1$.

The thresholding values for the SB_LACM model, the FP1_LACM model, and the FP2_LACM model are all chosen as $\gamma = 0.5$, which are used to find the segmented region $\Omega_1 = \{x : \phi(x) > \gamma\}$.

For synthetic and Envisat SAR images , we compare the *speed* of models ( the pair (.,.) is used to report both the number of iterations ( the first number ) and the CPU time) the second number )) , and *DSC* only for synthetic images, which are listed in TABLE Ⅰ, *pp* only for Envisat SAR images, which are listed in TABLE Ⅱ,respectively. All the models are implemented with Matlab 8.0 in core2 with 1.9 GHZ and 1GB RAM.

**A. Comparison LACM model with the CV model, the model [4] , and the RSF model**

Both the CV model and the model [4] all assume that the image intensity is piecewise constant and use the global intensity means to fit the image intensity. Therefore, they do not perform well in images with intensity inhomogeneity. Both the RSF model and the LACM model all assume that the image intensity is piecewise constant and use the local intensity means to fit the image intensity. The CV model, the model [4] and the RSF Model and the LACM model are all performed by level set method.

Although the RSF Model and the LACM model all use local means to fit the image intensity, the LACM model assumes that the image intensity is Gamma distribution with mean value $\mu_n = 1$ and variance $\sigma_n^2 = 1/L$, and the RSF Model assumes that the image intensity is Gaussian distribution. Therefore, the LACM model does perform well in images with intensity inhomogeneity than the RSF model.

In this section, we compare the LACM model with the CV model ,the model [4] and the RSF model by applying them to some real and synthetic images with intensity inhomogeneity in order to demonstrate the superior performance of our LACM to them for images with intensity inhomogeneity.

Fig. 1-Fig. 4 demonstrate the segmentation results on two synthetic images and two Envisat SAR images with intensity inhomogeneity by using the CV model, the model [4], RSF model and the LACM model, respectively.

As can seen from Fig.1(c)-(f) and Fig.2(c)-(f), i.e. the segmentation results of synthetic images, when the noise intensity is high (such as synthetic images 1), the CV model and the model [4] all result in severe misclassification, the RSF model gives a local minima, only can LACM model obtain a correct classification; When the noise intensity is low, the CV model and the RSF model give incorrect result, but the model [4] and the LACM model can give good results. Therefore, we can conclude that the LACM model yields satisfactory segmentation results because we consider and exploit the image local region information, which can better fit the image intensity distribution.

For Envisat SAR images, we also can see from Fig.3(b)-(d) and Fig.4(b)-(d) that the RSF model give a over segmentation result, the model in [4] can not segment the intensity inhomogeneity part of the SAR images, but the LACM model always give a correct result.

**B. Comparison the SB_LACM model with the model [5] and the model [26]**

In this section, we also use two synthetic image and two Envisat SAR images with severe intensity inhomogeneity to demonstrate the superior performance of globally convex based the SB_LACM model to the model [5] and the model [26].

Fig. 1-Fig. 4 demonstrate the segmentation results on two synthetic images and two real SAR images with intensity inhomogeneity by using the model [5], the model [26] , and the SB_LACM model and respectively.

We can see from Fig.1(g)-(i), Fig.2 (g)-(i), Fig.3(e)-(g) and Fig.4 (e)-(g)that SB_LACM model give the best result for the synthetic images and Envisat SAR images than the model [5], and the model [26].

Although the model [5] and the model [26] are all globally convex segmentation methods, they all can not fit the image intensity correctly because that the data term of the model [5] is based on the model same as the model [4] and the model [26] is based on the RSF model. However, the data term of our proposed Algorithm 1 is based on the LACM model, the SB_LACM model can give a good segmentation result for synthetic images and Envisat SAR images than the model [5] and the model [26].

In addition, we has transformed the proposed LACM model into a global optimization model by using convex relaxation technique, and apply the split Bregman technique to transform the global optimization model into two alternating optimization processes of Shrink operator and Laplace operator, so SB_LACM model only requires about 1/3 of the time needed for the LACM model (see TABLE Ⅰ, TABLE Ⅱ).

### C. Comparison the FP1_LACM model and the FP2_LACM model with the SB_LACM model

In this section, we also use the same two synthetic image and two Envisat SAR images with severe intensity inhomogeneity to demonstrate the superior performance of fixed point theory based the FP1_LACM model and the FP2_LACM model to the SB_LACM model.

Fig. 1-Fig. 4 demonstrate the segmentation results on two synthetic images and two real SAR images with intensity inhomogeneity by using the SB_LACM model, the FP1_LACM model and the FP2_LACM model, respectively.

We can see from Fig.1(i)-(k), Fig.2 (i)-(k), Fig.3 (g)-(i) and Fig.3 (g)-(i) that the FP1_LACM model and the FP2_LACM model all can give the best result for the synthetic images and Envisat SAR images than the SB_LACM model.

Because we has transformed the proposed LACM model into a global optimization model by using convex relaxation technique, and add the proximal term, and transform the proposed LACM model into a general classic ROF model by simple computing, then can obtain two fast algorithms for global optimization solver by using fixed point theory, which do not involve partial differential equations or difference equation, and only need simple difference compute, so the FP1_LACM model and the FP2_LACM model only require about 1/5 the time needed for the SB_LACM model (see TABLE Ⅰ, TABLE Ⅱ). They can further reduce the running time of the SB_LACM model needed.

**Remark:** By segmenting synthetic images and Envisat SAR images, the proposed algorithms not only can detect boundaries of images robustly and efficiently, and obtain a stationary global minimum, but also detect the images boundary more accurately.

**D. Quantitative Evaluation:**

We give the following several measures to evaluate the proposed model and Algorithms in this paper .

**1). Dice similarity coefficient**

The ground truth (GT) is drawn manually through visual inspection of the SAR images. The Dice Similarity coefficient (DSC) [27] between the computed segmentation (CS) and the GT is defined as:

$$DSC(CS,GT) = 2 \times \frac{N(CS \cap GT)}{N(CS) + N(GT)} \tag{36}$$

Where $N(\cdot)$ indicates the number of voxels in the enclosed set. The closer the *DSC* value to 1, the better the segmentation.

We use *DSC* value to evaluate segmentation accuracy of the LACM model and the the SB_LACM model , the FP1_LACM model and the FP2_LACM model on two synthetic images with different intensity inhomogeneity. The *DSC* values of the proposed LACM model and the SB_LACM model , the FP1_LACM model and the FP2_LACM model are shown in TABLE Ⅰ. As can be seen from TABLE Ⅰ, the *DSC* values of the proposed LACM model are higher than those of the model [4] and the RSF model. The *DSC* values of the SB_LACM model , the FP1_LACM model and the FP2_LACM model are higher than the proposed LACM model, the model [5] and the model [26].

**2) Uniformity measurement**

We adopt the uniformity measurement of image segmentation regions to evaluate the performance of the proposed method. The interior of each region should be uniform after the segmentation and there should be a great difference among different regions. That is to say, the uniformity degree of regions represents the quality of the segmentation. Therefore, we give the measurement of segmentation accuracy (SA) as follows [28]:

$$pp = 1 - \frac{1}{C}\sum_{i}\{\sum_{x\in R_i}[f(x) - \frac{1}{A_i}\sum_{x\in R_i} f(x)]^2\} \tag{37}$$

Where $R_i$ denotes different segmentation regions, $C$ is the normalization constant, $f(x)$ is the gray value of point $x$ in the image, $A_i$ is the number of the pixels in each region $R_i$. The closer to 1 the value of $PP$ is, the more uniform the interior of the segmentation regions are and the better the quality of the segmentation is.

We use *PP* value to evaluate segmentation accuracy of the LACM model and the SB_LACM model , the FP1_LACM model and the FP2_LACM model on two Envisat SAR images with different intensity inhomogeneity. As can be seen from TABLE Ⅱ, the *PP* values of the LACM model are higher than the model [4] and the RSF model, and the *PP* values of the SB_LACM model , the FP1_LACM model and the FP2_LACM model are all close to 1, which show high precision of the proposed model and Algorithms .

## Ⅴ.CONCLUSION

In this paper, we propose a LACM model, and apply convex relaxation technique to transform the proposed model into the global optimization model. Firstly, we apply the Split Bregman method to give an efficient SB-LACM model for SAR image segmentation , which can reduce greatly the running time of the LACM model needed.

In order to further reduce the running time of image segmentation, we propose two fast algorithm by applying different splitting approaches than the SB-LACM model to transform the proposed model to a general ROF model, which do not involve partial differential equations or difference equation, and only need simple difference compute. The FP1-LACM model or the FP2-LACM model can reduce 4/5 of the running time of the SB-LACM model needed. The FP1-LACM model or the FP2-LACM model can also be applied to the case of isotropic TV, and all the segmentation question are solved by the ROF model which has been intensively studied.

By segmenting synthetic images and Envisat SAR images, the proposed algorithms not only can detect boundaries of images robustly and efficiently, and obtain a stationary global minimum, but also detect the images boundary more accurately.

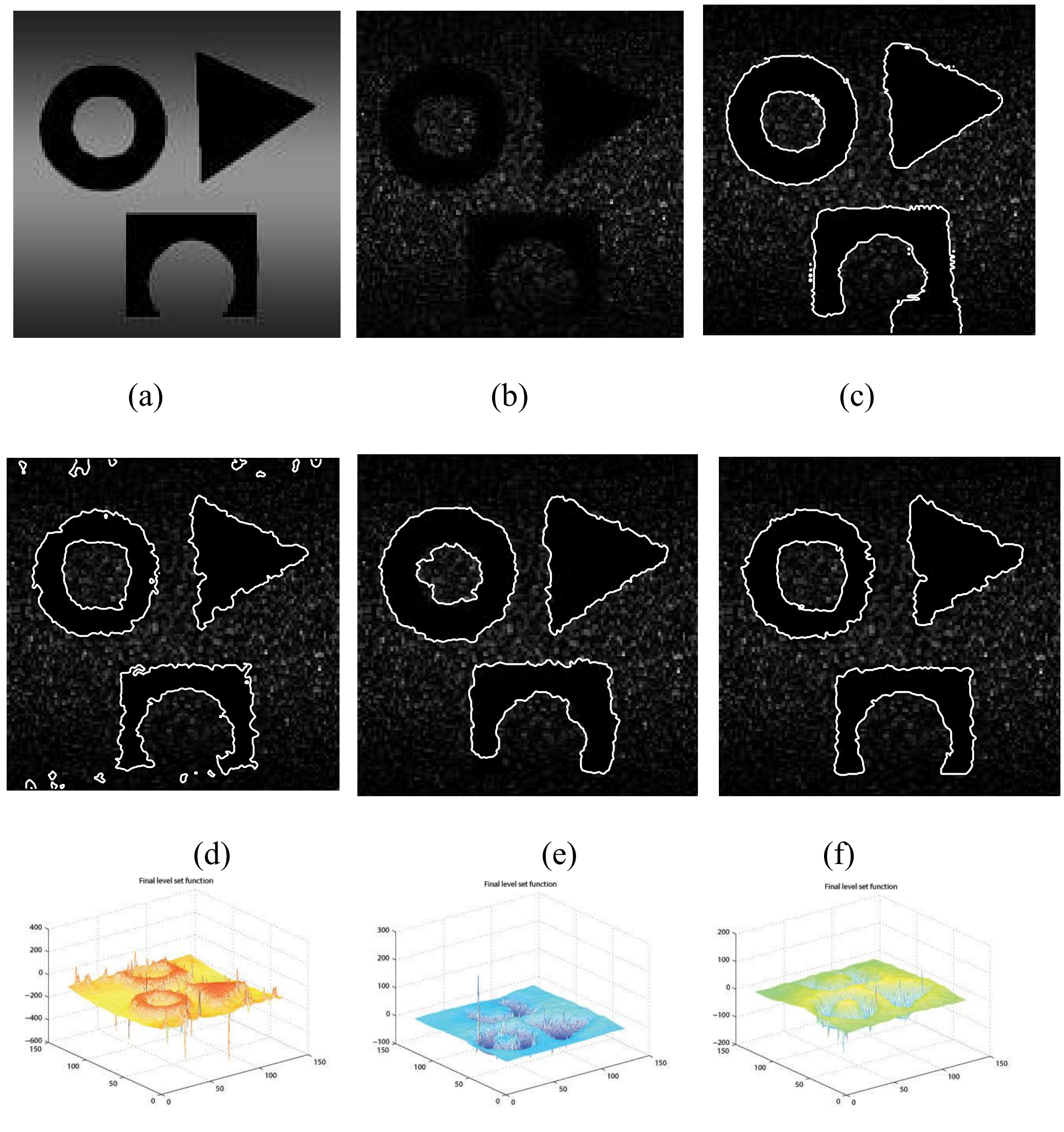

(a) (b) (c)

(d) (e) (f)

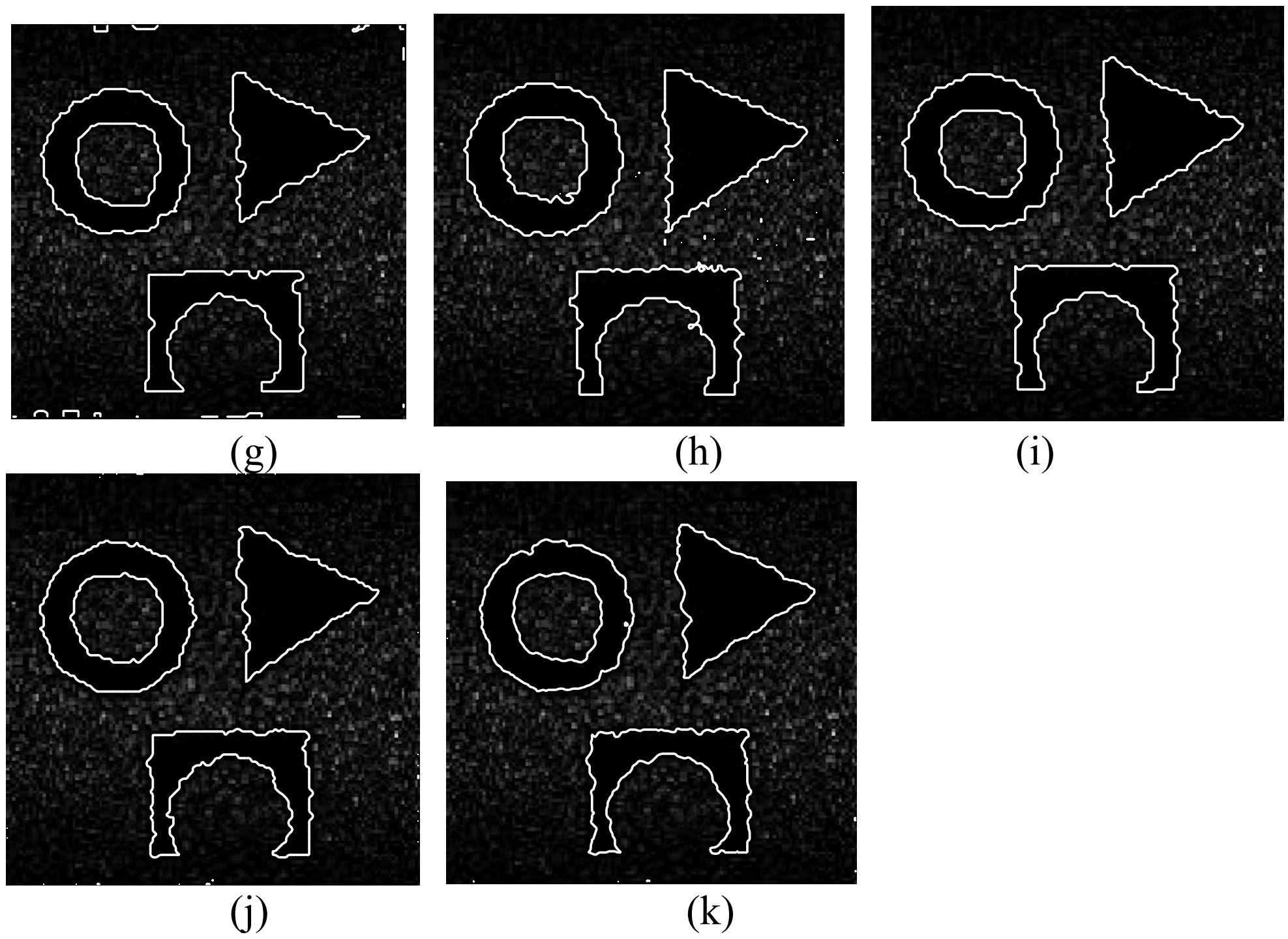

Fig.1. (a) original image.(b)synthetic image 1(*L*=1). (c) final contour by the CV model. **Level set method:** (d) final contour by the model [4]. (e) final contour by the RSF model. (f) final contour by the LACM model. **Split Bregman method:** (g) final contour by the model [5]. (h) final contour by the model [26]. (i) final contour by the SB_LACM model. **Fixed point method:** (j) final contour by the FP1_LACM model. (k) final contour by the FP2_LACM model.

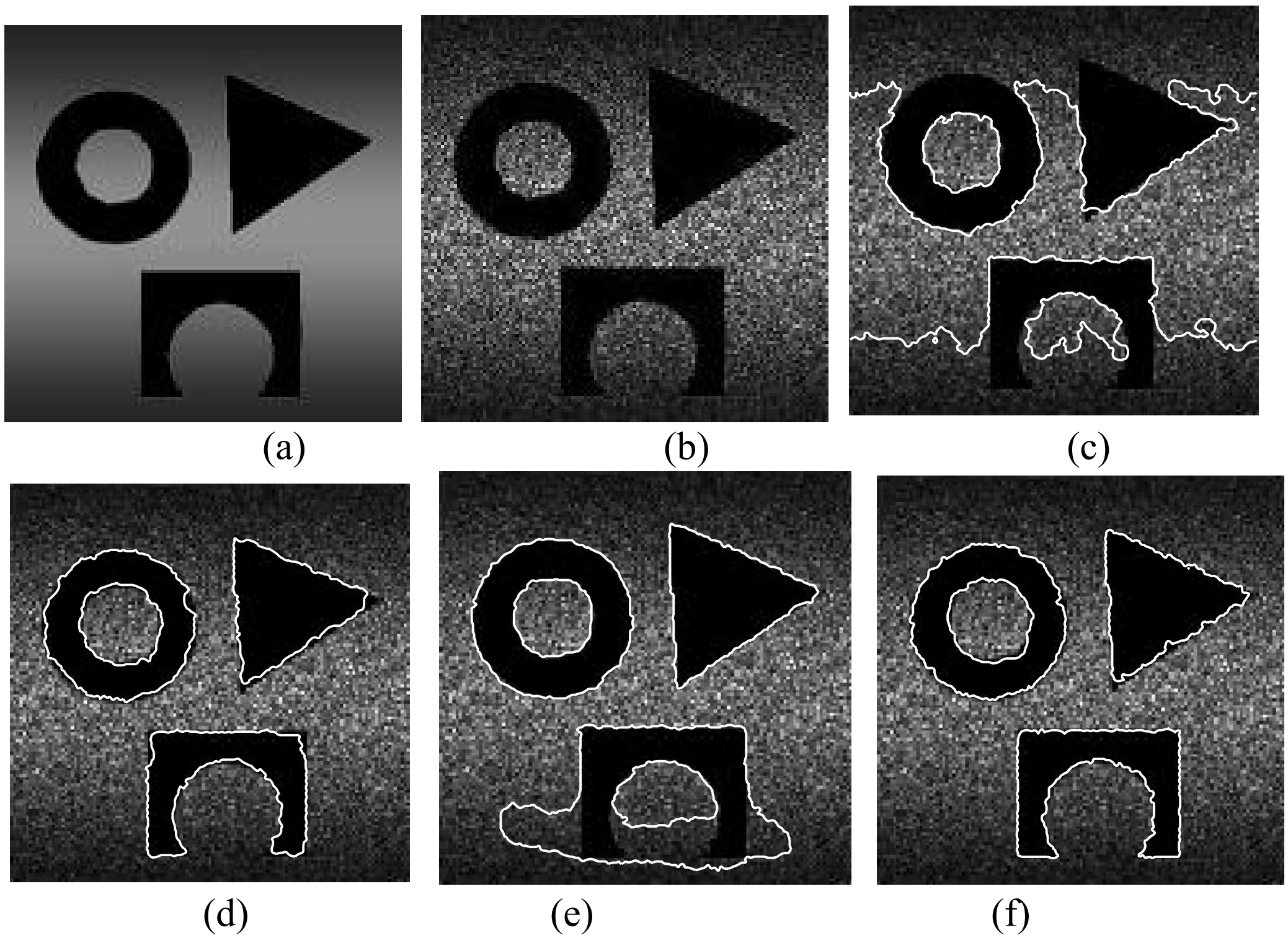

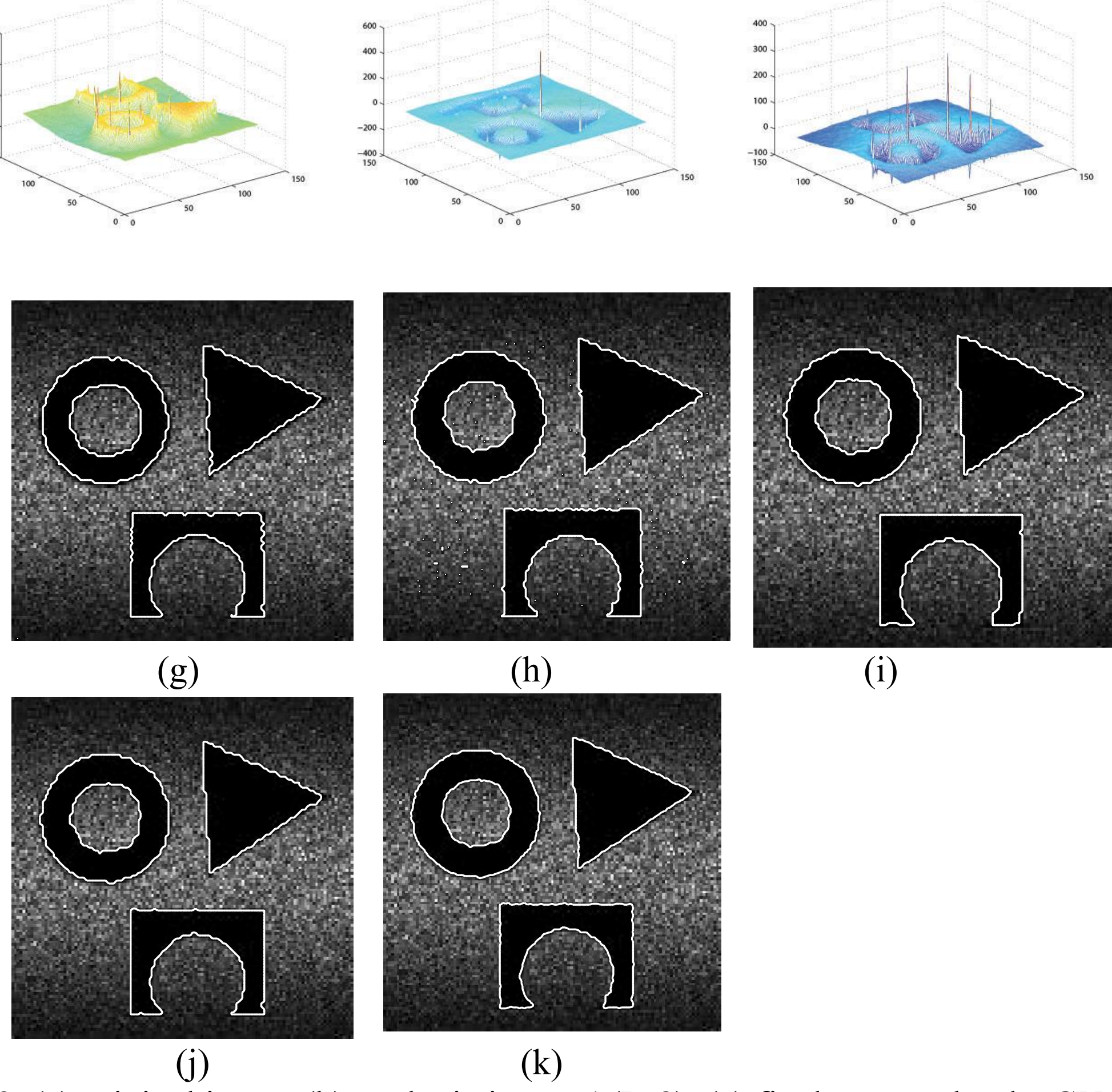

Fig.2. (a) original image.(b)synthetic image 1($L$=8). (c) final contour by the CV model. **Level set method:** (d) final contour by the model [4]. (e) final contour by the RSF model. (f) final contour by the LACM model. **Split Bregman method:** (g) final contour by the model [5]. (h) final contour by the model [26]. (i) final contour by the SB_LACM model. **Fixed point method:** (j) final contour by the FP1_LACM model. (k) final contour by the FP2_LACM model.

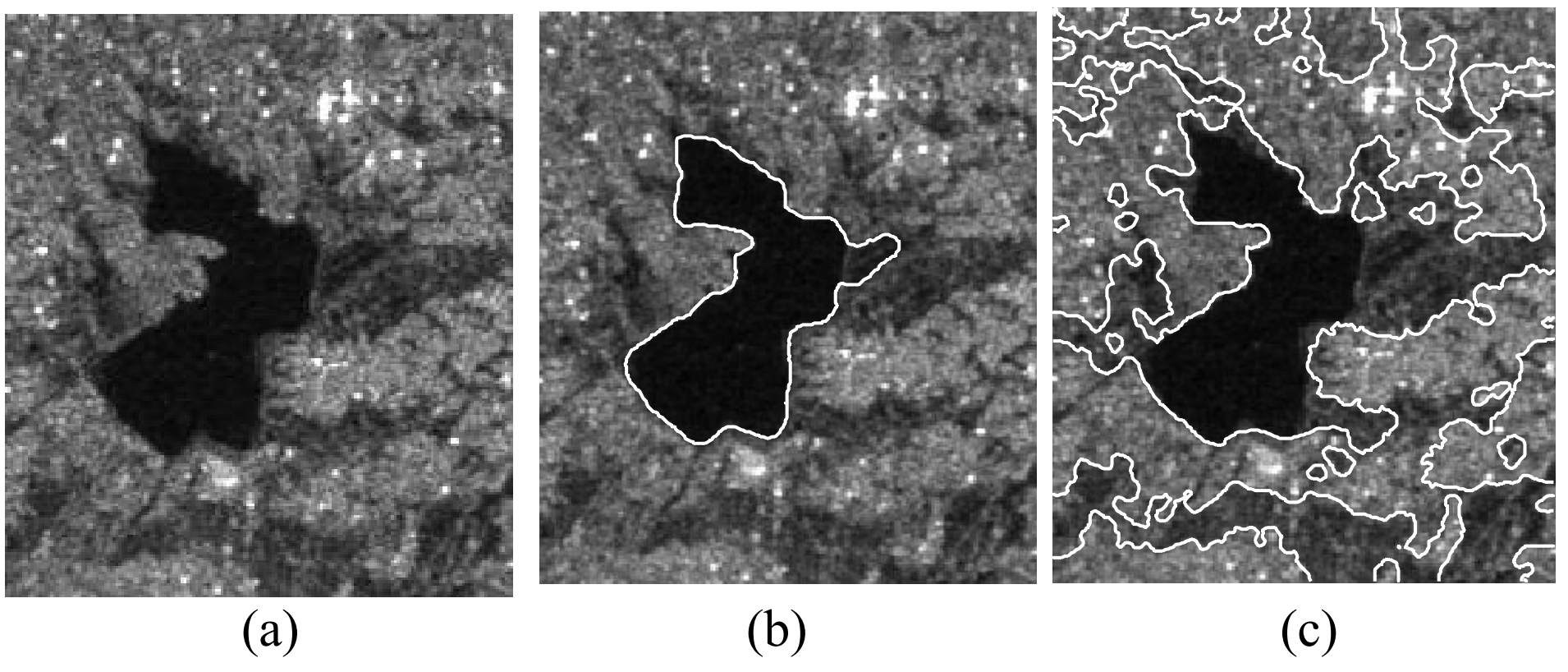

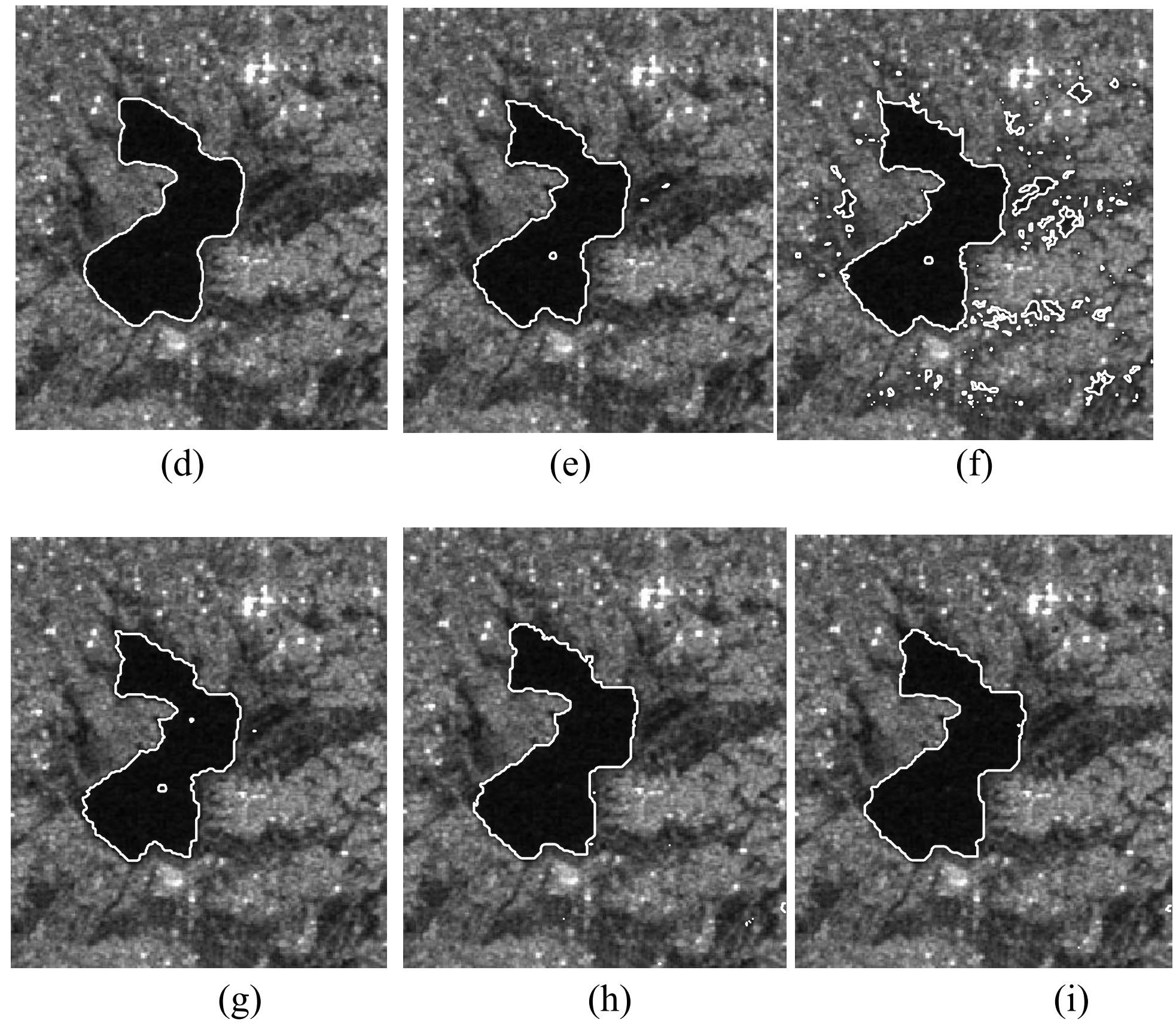

Fig.3.(a) SAR image 1. **Level set method:** (b) final contour by the model [4]. (c) final contour by the RSF model. (d) final contour by the LACM model. **Split Bregman method:** (e) final contour by the model [5]. (f) final contour by the model [26]. (g) final contour by the SB_LACM model. **Fixed point method:** (h) final contour by the FP1_LACM model. (i) final contour by the FP2_LACM model.

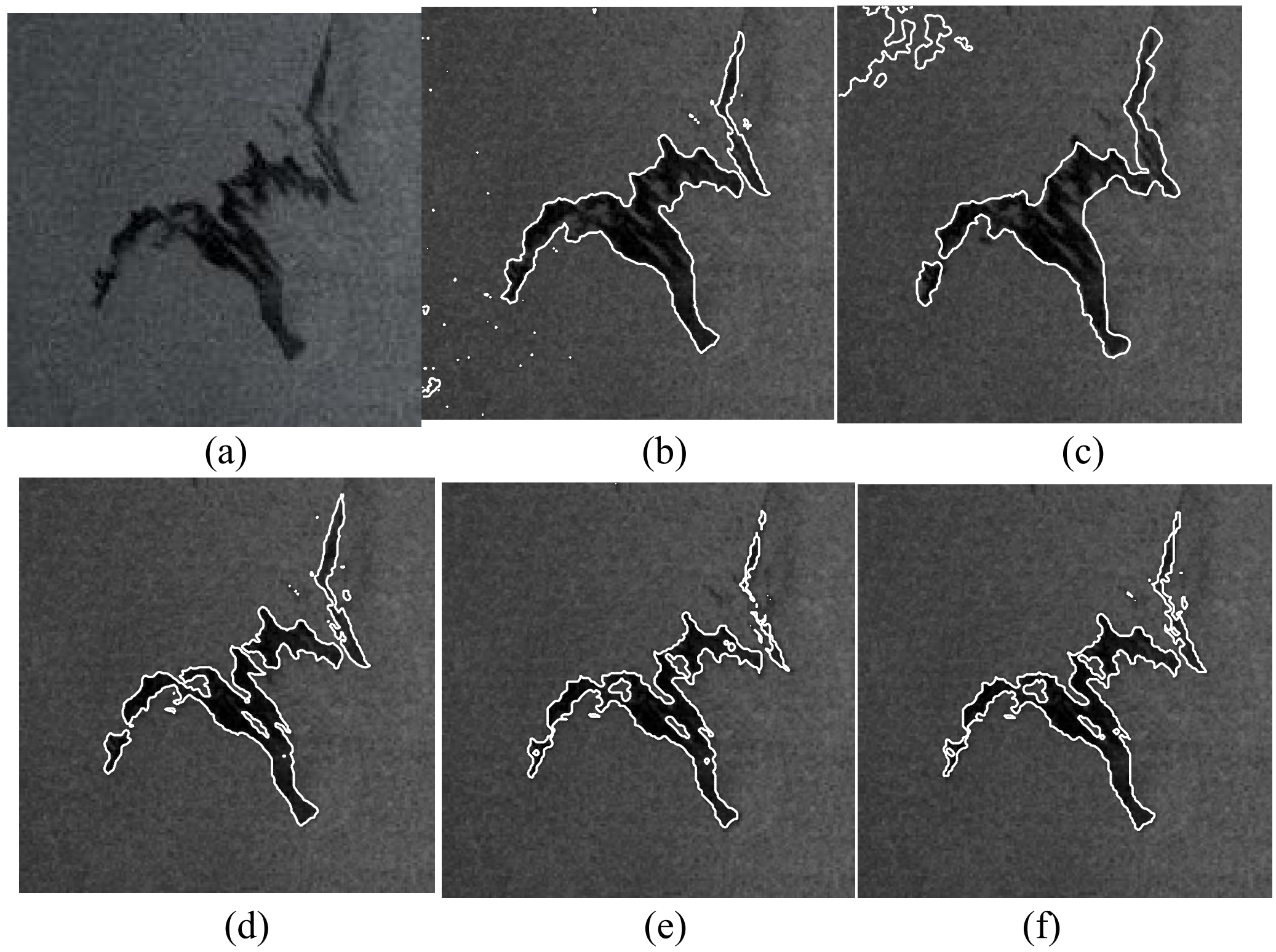

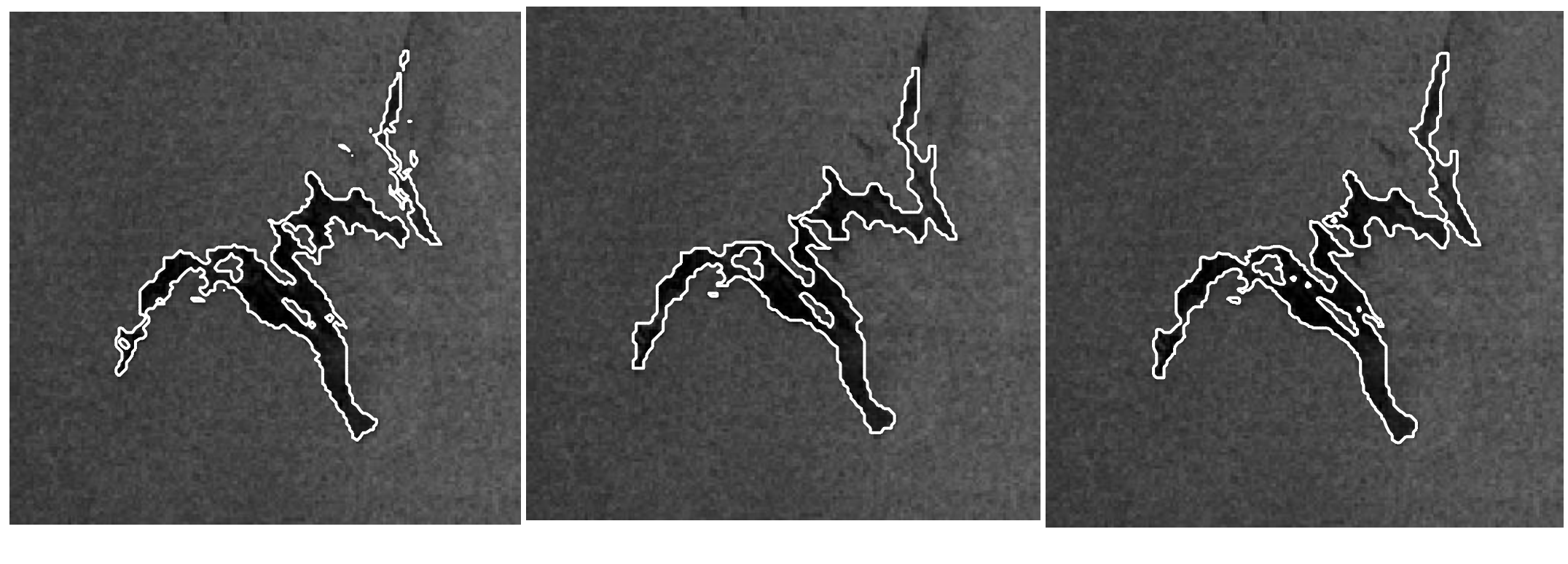

(g) (h) (i)

Fig.4.(a) SAR image 2. **Level set method:** (b) final contour by the model [4]. (c) final contour by the RSF model. (d) final contour by the LACM model. **Split Bregman method:** (e) final contour by the model [5]. (f) final contour by the model [26]. (g) final contour by the SB_LACM model. **Fixed point method:** (h) final contour by the FP1_LACM model. (i) final contour by the FP2_LACM model.

TABLE Ⅰ
PERFORMANCE OF OUR PROPOSED MODELS
AND OTHER MODELS

| models | synthetic image1 ($L$=1) | | synthetic image 2 ($L$=8) | |
|---|---|---|---|---|
| | ***Speed*** | ***DSC*** | ***Speed*** | ***DSC*** |
| **Level set method based (delta=15)** | | | | |
| model [4] | (200,12.203s) | 94.76% | (160,6.688s) | 86.52% |
| RSF model | (120,83.703 s) | 95.38% | (80,56.719s | 92.15% |
| LACM model | (100,68.563s) | 95.89% | (80, 56.172s) | 96.04% |
| **Split Bregman method based** | | | | |
| model [5] | (35,0.500s) | 95.16% | (20,0.328s) | 96.08% |
| model [26] | (30,20.891s) | 96.35% | (20,14.281s) | 96.27% |
| SB_LACM | (30,20.953s) | 96.51% | (20,14.36s) | 96.55% |
| **Fixed point method based (delta=15)** | | | | |
| FP1_LACM | (3,2.578s) | 96.63% | (3,2.641s) | 96.65% |
| FP2_LACM | (2,1.875s) | 96.94% | (2,1.906s) | 96.59% |

TABLE II

PERFORMANCE OF THE PROPOSED MODELS AND CLASSIC MODELS

| models | SAR image1 | | SAR image 2 | |
|---|---|---|---|---|
| | ***Speed*** | ***PP*** | ***Speed*** | ***PP*** |
| **Level set method based (delta=15)** | | | | |
| model [4] | (200,23.234s ) | 0.7821 | (200,21.453s) | 0.7265 |
| RSF mode | (150, 253.14s ) | -1.4459e+003 | (150,121.078s ) | -893.7310 |
| LACM model | (150,251.719s ) | 0.9486 | (100,81.296s ) | 0.9377 |
| **Split Bregman method based** | | | | |
| model [5] | ( 15,0.5s ) | 0.9990 | (15,0.343s ) | 0.9996 |
| model [26] | (15, 25.797s ) | 0.9994 | (15, 12.656s ) | 0.9996 |
| SB_LACM | (15,25.672s ) | 0.9994 | (25,20.078s ) | 0.9996 |
| **Fixed point method based (delta=8)** | | | | |
| FP1_LACM | (30,7.5s ) | 0.9994 | (50,7.625s) | 0.9996 |
| FP2_LACM | (30,7.547s ) | 0.9994 | (50,7.765s) | 0.9970 |